\relax
\documentclass[letterpaper]{article} 
\usepackage{aaai18}  
\usepackage{times}  
\usepackage{helvet}  
\usepackage{courier}  
\usepackage{url}  
\usepackage{epsfig}
\usepackage{graphicx}
\usepackage[fleqn]{amsmath}
\usepackage[ruled]{algorithm2e}
\usepackage{algorithmic}
\begin{document}

\title{EasiCS: the objective and fine-grained classification method of cervical spondylosis dysfunction}
\author{Nana Wang$^{1,2}$, Li Cui$^{1,4}$, Xi Huang$^{1}$, Yingcong Xiang$^{1,2}$, Jing Xiao$^{3}$, Yi Rao$^{3}$\\
$^{1}$Institute of Computing Technology(ICT), Chinese Academy of Sciences(CAS), Beijing, 100190, China.\\
$^{2}$ University of Chinese Academy of Sciences, Beijing, China.\\
$^{3}$Xiyuan Hospital, China Academy of Chinese Medical Sciences(CACMS), Beijing, China.\\
$^{4}$e-mail:lcui@ict.ac.cn\\
}
\maketitle

\begin{abstract}
The precise diagnosis is of great significance in developing precise treatment plans to restore neck function and reduce the burden posed by the cervical spondylosis (CS). However, the current available neck function assessment method are subjective and coarse-grained. In this paper, based on the relationship among CS, cervical structure, cervical vertebra function, and surface electromyography (sEMG), we seek to develop a clustering algorithms on the sEMG data set collected from the clinical environment and implement the division. We proposed and developed the framework EasiCS, which consists of dimension reduction, clustering algorithm EasiSOM, spectral clustering algorithm EasiSC. The EasiCS outperform the commonly used seven algorithms overall.
\end{abstract}

\section{Introduction}
The cervical spondylosis(CS), a common degenerative disease, harms human life and health, affects up to two-thirds of the population, and poses an serious burden on individuals and society~\cite{matz2009joint,kotil2008prospective,cai2016trend,wang2018,wang2018easicsdeep}.
Currently, the neck disability index~\cite{howardneck} is the most commonly used tool to assess the neck dysfunction~\cite{vernon1991neck},
The availability of which are mainly undermined by the coarse-grained and unreasonable classification, despite that the NDI information is subjective and not accurate enough.

The surface electromyography (sEMG) is a non-stationary weak physiological signal collected by the sEMG device, and consists of the Motor Unit Action Potential Trains (MUAPTs) which is generated by motor units and superimposed on the surface of the skin~\cite{wang2018}.
The sEMG is the CS-related physiological signals and have the ability to reflect the neck function status closely related to CS~\cite{wang2018easicsdeep,falla2007muscle,johnston2008alterations,johnston2008neck,madeleine2016effects}.
What's more, the signals is non-intrusive and affordable, and the acquisition is convenient.
Thus, we seek to use the sEMG data set collected from the clinical environment to provide more objective and fine-grained classification of cervical function.

As a powerful model-based clustering algorithm, the Self-organizing mapping (SOM) has strong ability of the self-learning, self-organizing, adaptive and nonlinear mapping, which is especially suitable for dealing with nonlinear reasoning, recognition, and classification task without the ground truth on the high dimensional and small-sampling data set~\cite{chenjunlin2017}.
Thus, we seek to utilize the SOM clustering algorithm to implement the division on the sEMG data set.
Despite of the advantages,
the stability is critical for clinical application research.
And, the stability clusters are defined in our paper as follow:
the stable differences between the individuals of different categories\footnote{the samples that are assigned to a cluster by a trained SOM model can also be assigned to the same cluster by second trained SOM.}.
However, it is an challenging task to obtain a stability clustering results as the samples that are assigned to a group by a trained SOM model are assigned to the different group by second trained SOM, compared with the clustering result of the two trained SOM with the different parameter settings.

In order to achieve it, we proposed and developed the classification framework EasiCS to obtain the relative stability clustering results, which consists of dimension reduction, clustering algorithm EasiSOM, spectral clustering algorithm EasiSC as shown in the Figure 1.
To the best of our knowledge, the EasiCS is the first effort to utilize the clustering algorithm and sEMG. Compared with the seven commonly used clustering algorithms, the novelty framework EasiCS provide the best overall performance.
The cervical spondylosis(CS), a common degenerative disease, harms human life and health, affects up to two-thirds of the population, and poses an serious burden on individuals and society~\cite{matz2009joint,kotil2008prospective,cai2016trend,wang2018,wang2018easicsdeep}.
Currently, the neck disability index~\cite{howardneck} is the most commonly used tool to assess the neck dysfunction~\cite{vernon1991neck},
The availability of which are mainly undermined by the coarse-grained and unreasonable classification, despite that the NDI information is subjective and not accurate enough.

The surface electromyography (sEMG) is a non-stationary weak physiological signal collected by the sEMG device, and consists of the Motor Unit Action Potential Trains (MUAPTs) which is generated by motor units and superimposed on the surface of the skin~\cite{wang2018}.
The sEMG is the CS-related physiological signals and have the ability to reflect the neck function status closely related to CS~\cite{wang2018easicsdeep,falla2007muscle,johnston2008alterations,johnston2008neck,madeleine2016effects}.
What's more, the signals is non-intrusive and affordable, and the acquisition is convenient.
Thus, we seek to use the sEMG data set collected from the clinical environment to provide more objective and fine-grained classification of cervical function.

As a powerful model-based clustering algorithm, the Self-organizing mapping (SOM) has strong ability of the self-learning, self-organizing, adaptive and nonlinear mapping, which is especially suitable for dealing with nonlinear reasoning, recognition, and classification task without the ground truth on the high dimensional and small-sampling data set~\cite{chenjunlin2017}.
Thus, we seek to utilize the SOM clustering algorithm to implement the division on the sEMG data set.
Despite of the advantages,
the stability is critical for clinical application research.
And, the stability clusters are defined in our paper as follow:
the stable differences between the individuals of different categories.
However, it is an challenging task to obtain a stability clustering results as the samples that are assigned to a group by a trained SOM model are assigned to the different group by second trained SOM, compared with the clustering result of the two trained SOM with the different parameter settings.

In order to achieve it, we proposed and developed the classification framework EasiCS to obtain the relative stability clustering results, which consists of dimension reduction, clustering algorithm EasiSOM, spectral clustering algorithm EasiSC as shown in the Figure 1.
To the best of our knowledge, the EasiCS is the first effort to utilize the clustering algorithm and sEMG. Compared with the seven commonly used clustering algorithms, the novelty framework EasiCS provide the best overall performance.

\begin{center}
\begin{figure}
  \epsfig{file=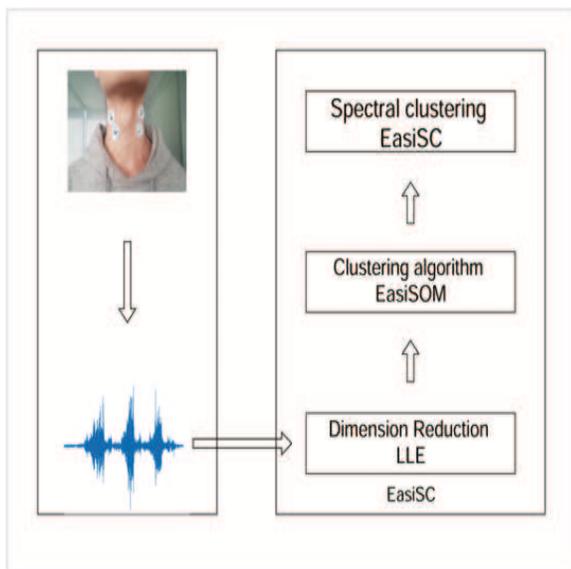, width = 3.00in, height = 3.00in}
  \caption{The EasiCS. \label{fig: Figure 5}}
\end{figure}
\end{center}

The CS is a chronic musculoskeletal disorder, which is mainly accompanied by the neck pain and the disability of human-related functions~\cite{hogg2009burden}.
The current classification of the CS is based on clinical symptoms and cervical lesion which was in accordance with 2012 ICD-9-CM Diagnosis Code 721(721.0 Cervical spondylosis
without myelopathy, 721.1 Cervical spondylosis with myelopathy) and the diagnostic criteria
of diagnosis and treatment for CS issued by China Rehabilitation Medicine Association~\cite{guidecs2010}.
There are many studies on CS intelligent classification: clinical-symptoms-based method, cervical-vertebra -lesion-based method, neck-muscle-lesion-based method, traditional Chinese medical classification.

For the clinical-symptoms-based method, with the guidance of the knowledge engineering and expert system construction theory, the work~\cite{jebri2015detection} developed the medical diagnostic expert system CSES of cervical spondylosis which use the forward reasoning as the main reasoning mechanism and production rules to represent domain expert knowledge.

For the cervical-spine-lesion-based method, faced with the divergences in the traditional X-ray reading method, the research~\cite{yu2015classifying} proposed the method based on maximum likelihood theory to solve the type classification of CS, and it is proved to be the effective method.
For the degenerative changes of the cervical spine, the research~\cite{jebri2015detection} proposed a machine learning approach to detect and localize degenerative changes in lateral X-ray images of the cervical spine, obtaining the 95\% accuracy.

For the neck-muscle-lesion-based method, the work~\cite{CS2011zhaozhongmin} have demonstrated that there are the stability and reproducibility of myoelectric activity on the surface of normal human cervical muscles, and there are significant differences of surface electromyography index between the cervical spondylosis and the healthy.
The work~\cite{wang2018} proposed a convenient non-harm CS intelligent identify method EasiCNCSII which consists of the sEMG data acquisition and the CS identification, obtaining the best performance of 91.02\% in mean accuracy, 97.14\% in mean sensitivity, 81.43\% in mean specificity, 0.95 in mean AUC. The research~\cite{wang2018easicsdeep} proposed an intelligent method EasiDeep based on the deep learning which utilized the surface electromyography (sEMG) signal to identify CS, achieving the state-of-the-art performance.
It proves that sEMG contains pathological information of CS which is consistent with the
work~\cite{falla2007muscle,johnston2008alterations,johnston2008neck,madeleine2016effects}.

The application of clustering technology has achieved highlighted results in the bioinformatics field as well as the field of image segmentation, object and character recognition.
Faced with the problem that the extent to which genomic signatures
are shared across tissues is still unclear, the Katherine~\cite{hoadley2014multiplatform} developed an integrative analysis to reveal a unified classification into 11 major subtypes.
The research~\cite{shen2009integrative} developed a joint latent variable model iCluster for integrative
clustering to analyze breast and lung cancer subtype.
The research~\cite{shen2012integrative} utilized the iCluster to present an integrative subtype analysis of the TCGA glioblastoma (GBM) data set, revealing new insights through integrated subtype characterization.

\section{Preliminaries}
\subsection{Participants}

The 57 volunteers participated in the study from March 15, 2017 to July 15,
2018 in China, the female number of which is 42 and the male number of which is 15.
The subjects have received a clinical diagnosis of the CS, which are in accordance with 2012 ICD-9-CM Diagnosis Code 721 (721.0 Cervical spondylosis without myelopathy, 721.1 Cervical spondylosis with myelopathy) and the criteria of diagnosis and treatment for CS issued by China Rehabilitation Medicine association. The 57 subjects are mainly sedentary people from 27 different occupations and involve 27 types of the CS lesion, the age of which range from 20 to 64.

\subsection{Dataset}

The data set are acquired from 57 volunteers above. The sEMG signal were synchronously recorded from the 6 muscles: the left sternocleidomastoid ($M0$), the left upper trapezius ($M1$), the left cervical erector spinae ($M2$), the right cervical erector spinae ($M3$), the right upper trapezius ($M4$) and the right sternocleidomastoid ($M5$).
The volunteers complete the 7 movements $Aj$ ($0 < j < 7$) in sequence, each movement of which is performed 3 times.
The sEMG data $S_{i,j}$ is obtained from the muscle $M_i$ activated by the movement $A_j$.
The $2949$ feature are extracted from the $S_{i,j}$ ($0 < i < 6, 0 < j < 7$), forming a sample $S$.
The detail on the $S$ is in the paper~\cite{wang2018}.
The 3 samples are from a volunteer, and the 171 ($57 \times 3$) samples is obtained from the 57 volunteers.

\section{Methodology}

In this section, we elaborated our proposed method EasiCS.
As shown in Figure 2, the EasiCS consists of three parts: the dimensionality reduction, clustering algorithm EasiSOM, community detection EasiSC.

\begin{center}
\begin{figure}
\centering
\epsfig{file = 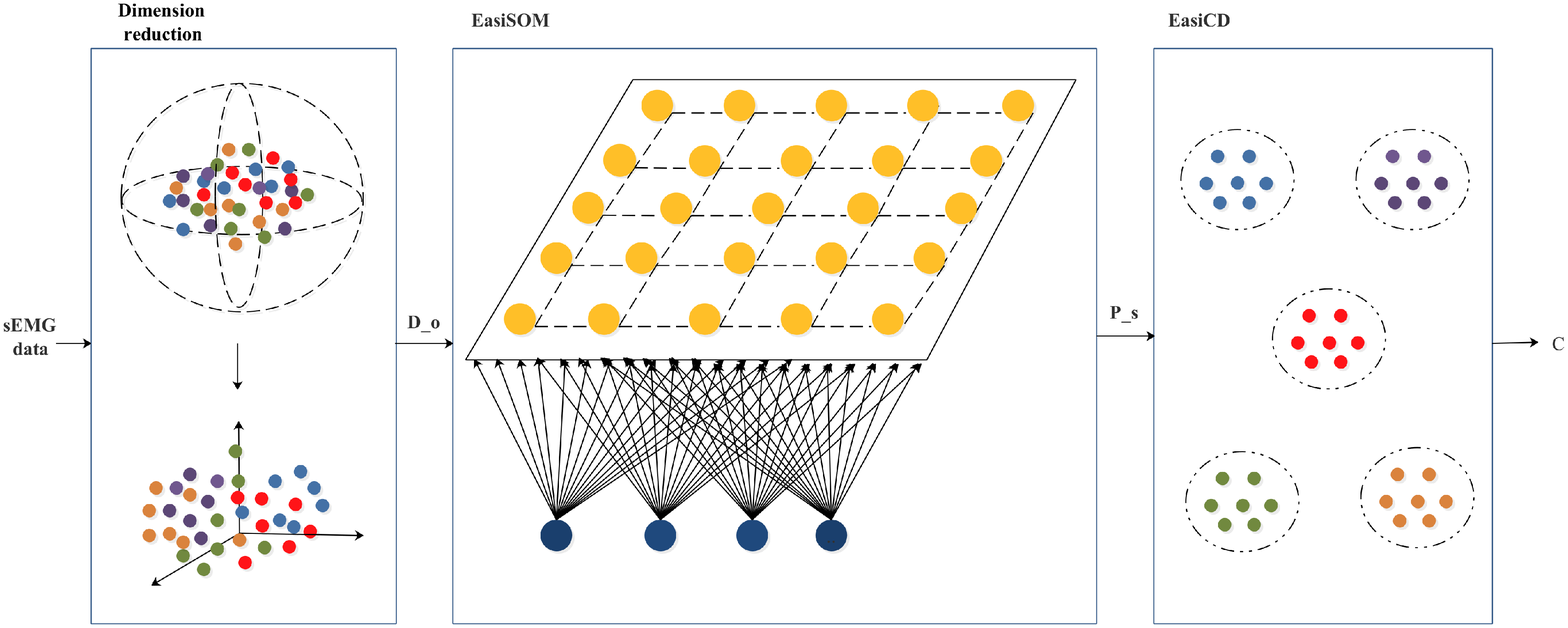, width = 3.5in, height = 2.00in}
\caption{The EasiCS.}
\label{fig: Figure 6}
\end{figure}
\end{center}

\subsection{The dimensionality reduction: locally linear embedding}

The high-dimensional sEMG data decreases computational efficiency, increase storage overhead, and cause overfit~\cite{li2018feature,wang2018}, especially for the small sample data sets.
Faced with the high-dimensional data, dimensionality reduction is an effective means of data preprocessing.
Unlike clustering methods for local dimensionality reduction, the locally linear embedding (LLE), an unsupervised learning algorithm, computes low-dimensional, neighbor-hood-preserving embeddings of high-dimensional inputs and is able to learn the global structure of nonlinear manifolds~\cite{roweis2000nonlinear}.

In this paper, we utilized the LLE to deal with the input according to the work\cite{roweis2000nonlinear}.
The number of neighbors and the dimension of the data set is set to 30.
The $D_o$ is obtained.
The general process is shown in the Algorithm \ref{alg:alg1}.
The details on calculation are in work~\cite{roweis2000nonlinear}.

\begin{align}\label{eq8}
Z_i &= (X_i - X_j)^\mathrm{T}(X_i - X_j)\\
\hspace{0.8cm}
W_i &= \frac {Z_i^{-1}1_k} {1_k^T Z_i^{-1} 1_k}\\
\hspace{0.8cm}
M &= (1 - W)(1 - W)^T
\end{align}

\subsection{The EasiSOM}

The Self-Organizing Map(SOM) is an excellent tool in exploratory phase of data mining, and projects input space on prototypes of a low-dimensional regular grid that can be effectively utilized to visualize and explore properties of the data~\cite{vesanto2000clustering}.

The SOM clustering results are sensitive to initial value settings, leading that there are different
clusters result between two trained SOM model with the different parameter settings.
And the input $D_o$ of the SOM contains 171 samples from the 57 subjects.
A division of the 171 samples reflect the internal relations of the 57 subjects above.
Although the divisions are different, we think they contain the common information that reflect the true association of 57 subjects. Thus, according to the ensemble learning, we utilize the multiple SOM algorithm to perform multiple classifications and find a set of partitions to analyze the internal commonality of the division above.

In this paper, as shown in the Algorithm \ref{alg:alg1}, we developed the EasiSOM based on the SOM in our paper, which integrate the 1000 trained SOM to generate the 1000 division results.
Finally, a partition set $P_s$ is obtained which includes 625 divisions.

\begin{algorithm}
    \renewcommand{\algorithmicrequire}{\textbf{Input:}}
    \renewcommand{\algorithmicensure}{\textbf{Output:}}
    \caption{The EasiSOM algorithm}
    \label{alg:alg1}
    \begin{algorithmic}[1]
    \REQUIRE the data set $D_o$, the partition number $n_{p}$, the number $In_{n}$ of input neurons, the number $Out_{n}$ of output neurons, weight vector $W$, learning rate $l$, learning rate threshold $l_t$, neighborhood size $r$, and the maximum number $iter_{m}$ of iterations.
    \ENSURE The $P_s$ including the $iter$ division of the data set $D_o$.
    \STATE Initialize the iteration number $iter = 0$, $In_{n}$, \\
           $Out_{n}$, $iter_{m} = 10000$ and $n_{p} = 1000$;
    \FOR{$d_n = 0$ to $d_n = n_{p}$}
    \WHILE{$iter < iter_{m}$ or $l < l_t$}
    \STATE Initialize the $W$, $l$, $l_t$, $r$.
    \STATE Calculate the winning node for each input sample $X_i$;
    \STATE Update the weight vector $W$, learning rate $l$;
    \STATE $iter = iter + 1$
    \ENDWHILE
    \STATE Compute the division $p_{iter}$ of the data set $D_o$;
    \IF{$ICS < 0.099$}
    \STATE Add the partition $p_{iter}$ to the set $P_s$.
    \ENDIF
    \ENDFOR
    \end{algorithmic}
\end{algorithm}

\subsection{The EasiSC}

Next, we seek to find the internal structure of 57 entities behinds the $P_s$.
The network is a powerful mechanism, and has the ability to represent the complex relationship between the data~\cite{huang2018clustering}. 
Complex systems with interconnected internal entities can be abstracted into networks which have extremely important structure information. 
The identification of the structure is of crucial importance as they may help to uncover a-prior unknown functional module such as topics in information network or cyber-communities in social networks~\cite{blondel2008fast}.
The spectral clustering algorithm is a classic clustering algorithm based on network topology.

In our work, the EasiSC based on the spectral clustering algorithm are developed to cluster the $P_s$.
As shown in the Algorithm \ref{alg2}, the each sample $i$ is treated as the network node $N_i$,
and the edge between the node $N_i$ and $N_j$ is represented as $e_{i,j}$.
The weight value between nodes is initialized to 0.
If the two nodes are divided into the same class in the $k-th$ division, the weight value is increased by 1. The detail computation on the $W_{i,j}$ is shown in the Formula ~\ref{eq7}.

In order to facilitate the subsequent data analysis of 57 subjects, as the clustering results of the three samples of almost all subjects are the same,
we mapped the clustering result labels of 171 samples to 57 subject as following rule:
the label of a subject is defined as the labels that most samples of a subject have.

\begin{equation}\label{eq7}
\centering
W_{i,j}\!= \sum_{i=1}^{k}{X(i, j)}
\end{equation}

\begin{small}
\[X(i, j)=
\begin{cases}
1 &  \text{Sample $i$, $j$ are divided into the same class.} \\
0 &  \text{Sample $i$, $j$ are divided into the different class.}
\end{cases}
\]
\end{small}

Where the $k$ is the number of the divisions from the 625 trained SOM, and the $i$ and $j$ is the number of the 171 samples.

\begin{algorithm}
    \renewcommand{\algorithmicrequire}{\textbf{Input:}}
    \renewcommand{\algorithmicensure}{\textbf{Output:}}
    \caption{The EasiSC algorithm}
    \label{alg2}
    \begin{algorithmic}[1]
    \REQUIRE $P_s$
    \ENSURE  The partition $C$ of the $P_s$;
    \STATE   Computer the minimum cluster number $min$;
    \STATE   The maximum cluster number $max$;
    \STATE   Generate weight matrix $W$;
    \STATE   $k = min$
    \WHILE   {$k < max$}
    \STATE   Construct the similar matrix $S$ = $W$;
    \STATE   Construct the adjacent matrix $A$ = $S$;
    \STATE   Construct a standardized Laplacian matrix $L_m$;
    \STATE   Calculate $k$ minimum eigenvalues and eigenvectors;
    \STATE   Use eigenvectors to transform data set $P_{id}$ to $k$ dimension $P_{idk}$;
    \STATE   Use the k-means to cluster the data $P_{idk}$ obtain the $C$.
    \STATE   Compute the $SC$;
    \STATE   Add $SC$ to $SC\_list$;
    \STATE   Add $i$ to $i\_list$;
    \STATE   Add $C_{i}$ to $C\_list$;
    \STATE   $k = k+1$
    \ENDWHILE
    \STATE Calculate the partition $C_k$ with the largest $SC$.
    \STATE Map the clustering result $C_k$ of 171 samples to 57 subject.
    \end{algorithmic}
\end{algorithm}

\subsection{Algorithm}

As the Algorithm \ref{alg3} shown, we firstly reduce the dimensionality of the sEMG data $D$ and generate the data set $D_o$. Then, the EasiSOM is trained to iteratively cluster the $D_o$ and generate the $P_s$.
Finally, the EasiSC is trained to cluster $P_s$.

\begin{algorithm}
    \renewcommand{\algorithmicrequire}{\textbf{Input:}}
    \renewcommand{\algorithmicensure}{\textbf{Output:}}
    \caption{The EasiCS algorithm}
    \label{alg3}
    \begin{algorithmic}[1]
    \REQUIRE sEMG data set $D$
    \ENSURE the partition $C$
    \STATE Use LLE to reduct the dimension of $D$;
    \STATE Use EasiSOM to cluster $D_o$;
    \STATE Use EasiSC to cluster the $P_s$.
    \end{algorithmic}
\end{algorithm}

\section{Experiments Result and Discussion}

\subsection{The metrics}

For the lack of the ground truth (the truth label), it is generally believed that the optimal clustering partition minimizes the intra-cluster distance and maximizes inter-cluster distance.
The metric~\cite{metric} of the Silhouette Coefficient ($SC$), Calinski Harabaz score ($CH$), Davies Bouldin score ($DB$) are the widely used verification indicator to evaluate the quality of the clusters or the performance of the clustering algorithm.
The large Silhouette Coefficient is, the large Calinski Harabaz score is, the smaller Davies Bouldin score is, the better the quality of the clusters are, the better the clustering algorithm perform.
Thus, we utilized the indicators $SC$, $CH$ and $DB$ for overall cluster quality evaluation.

\subsection{The Model for comparison}

We compared the EasiSOM with the seven commonly used clustering models: k\-means, affinity propagation, Mean\-shift, spectral clustering, agglomerative clustering, gaussian mixtures and Birch on the same data set in the same clustering task as follows:

\begin{itemize}
\item K\-means (\textbf{$M_0$}): The K-Means is an algorithm for clustering data points by computing the average value. It is one of the best-known, bench marked and simplest clustering algorithms~\cite{macqueen1967some,saxena2017review}.

\item Affinity propagation (\textbf{$M_1$}): Affinity propagation~\cite{frey2007clustering} is a clustering algorithm based on 'information transfer' between data points, which does not need to specify the number of clusters in advance, and can automatically generate the optimal number of clusters.

\item Mean\-shift (\textbf{$M_2$}): it is a general nonparametric technique, which is used for the analysis of a complex multimodal feature space and to delineate arbitrarily shaped clusters~\cite{comaniciu2002mean}.

\item Spectral clustering (\textbf{$M_3$}): it is a graph-based clustering method. It is simple to implement, can be solved efficiently by standard linear algebra method, and outperforms traditional clustering algorithms~\cite{von2007tutorial}.

\item Agglomerative clustering (\textbf{$M_4$}): it is an hierarchical clustering methods and follows the bottom-up approach~\cite{saxena2017review}.

\item Gaussian mixtures (\textbf{$M_5$}): it is a parametric probability density function and commonly used as a parametric model of the probability distribution of continuous measurements or features in a biometric system~\cite{reynolds2015gaussian}.

\item Birch (\textbf{$M_6$}): it is a typical integrated hierarchical clustering algorithm~\cite{zhang1996birch,Ding2015Research}, which is more suitable for the large amount of data and the large number of clusters.
\end{itemize}

\subsection{The comparison of the performance of the model}

\begin{table}[ht]
\centering
\caption{The performance comparison between different clustering algorithms.}
\label{tab:tab4}
\begin{tabular}{cccccccc}
\hline
\hline
\\
&N &SC &CH &DB  &ICS\\
\\
\hline
\\
$M_0$            &56             &0.4871    &8.7952     &0.9686  &1.75\%\\
\\
$M_1$            &46             &0.4547    &7.8716     &1.1452  &2.34\%\\
\\
$M_2$            &7              &0.2331    &7.2309     &1.1274  &2.92\%\\
\\
$M_3$            &48             &0.4226    &7.5257     &1.2397  &7.60\%\\
\\
$M_4$            &56             &0.4535    &8.7116     &0.9800  &5.26\%\\
\\
$M_5$            &55             &0.4848    &8.7501     &0.9808  &3.51\%\\
\\
$M_6$            &56             &0.4803    &8.9374     &0.9687  &4.68\%\\
\\
SOM              &13             &0.7551    &2740.1569  &0.3399  &4.09\%\\
\\
\hline
\\
\textbf{EasiCS} &\textbf{5} &\textbf{0.8220} &\textbf{823.1703} &\textbf{0.2408} &\textbf{1.17\%}\\
\\
\hline
\hline
\end{tabular}
\end{table}

In order to evaluate the effectiveness of the clustering algorithm, with the metric of $SC$, $CH$, $DB$, and $ICS$, we compared the performance of the EasiCS with the seven commonly used models as shown in the table \ref{tab:tab4}.
And, the EasiCS obtained the largest $SC$ 0.8220, the second largest value $CH$ 823.1703, the smallest $DB$ 0.2408, outperforming the seven algorithms overall.

\subsubsection{The model consistency comparison}

In order to evaluate the stable of the clustering algorithm, with the metric of $Std\_num$, $Std\_SC$, $Std\_CH$, $Std\_DB$ and $Std\_ICS$, we compared the performance EasiCS with the SOM in the table \ref{tab:tab4_2}.
And, the EasiCS obtained the smallest $Std\_num$ 1.2635, the smallest $Std\_SC$ 0.1188, the smallest $Std\_CH$ 606.3688, and the smallest $Std\_DB$ 0.0989.
Compared with the SOM, the EasiCS have the smaller change, which means that the EasiSOM have the more stable cluster result.

\begin{table}
\caption{The consistency comparison.}
\label{tab:tab4_2}
\centering
\begin{tabular}{ccccc}
\hline
\hline
\\
         &$Std\_SC$          &$Std\_CH$         &$Std\_DB$  \\
\\
\hline
\\
$SOM$      &0.1233           &5982.5545        &0.1696   \\
\\
$EasiCS$   &0.1188           &606.3688         &0.0989   \\
\\
\hline
\hline
\end{tabular}
\end{table}

\section{Conclusions}

In this study, based on the relationship between cervical structure, cervical vertebrae function and CS, we developed the novelty framework based on the clustering algorithms to implement the division of the function, using the high-dimensional and small-sampling of the CS-related sEMG data,
which consists of the dimension reduction, clustering algorithm EasiSOM, spectral clustering algorithm EasiSC.
With the metric of the Silhouette Coefficient, the Calinski Harabaz score and the Davies Bouldin score, the EasiCS achieved the best overall performance compared with the seven commonly used clustering algorithms.

Thus, we will collect and utilize more diverse and comprehensive data to classify CS more precisely and explore the law behind the pathogenesis of CS and classification, providing more knowledge about CS prevention and treatment.


\begin{thebibliography}{10}

\bibitem[\protect\citeauthoryear{Blondel \bgroup et al\mbox.\egroup
  }{2008}]{blondel2008fast}
Blondel, V.~D.; Guillaume, J.-L.; Lambiotte, R.; and Lefebvre, E.
\newblock 2008.
\newblock Fast unfolding of communities in large networks.
\newblock {\em Journal of statistical mechanics: theory and experiment}
  2008(10):P10008.

\bibitem[\protect\citeauthoryear{Cai \bgroup et al\mbox.\egroup
  }{2016}]{cai2016trend}
Cai, Z.; Zhang, N.; Ma, N.; Dong, G.; Wang, S.; and Zhao, Y.
\newblock 2016.
\newblock Trend of the incidence of cervical spondylosis: decrease with aging
  in the elderly and increase with aging in the young and the adults.
\newblock {\em Int J Clin Exp Med} 9(7):14329--14336.

\bibitem[\protect\citeauthoryear{Comaniciu and Meer}{2002}]{comaniciu2002mean}
Comaniciu, D., and Meer, P.
\newblock 2002.
\newblock Mean shift: A robust approach toward feature space analysis.
\newblock {\em IEEE Transactions on Pattern Analysis \& Machine Intelligence}
  (5):603--619.

\bibitem[\protect\citeauthoryear{Ding \bgroup et al\mbox.\egroup
  }{2015}]{Ding2015Research}
Ding, S.; Wu, F.; Qian, J.; Jia, H.; and Jin, F.
\newblock 2015.
\newblock Research on data stream clustering algorithms.
\newblock {\em Artificial Intelligence Review} 43(4):593--600.

\bibitem[\protect\citeauthoryear{Falla \bgroup et al\mbox.\egroup
  }{2007}]{falla2007muscle}
Falla, D., .; Farina, D., .; M~Kanstrup, D.; and Graven-Nielsen, T., .
\newblock 2007.
\newblock Muscle pain induces task-dependent changes in cervical
  agonist/antagonist activity.
\newblock {\em Journal of Applied Physiology} 102(2):601.

\bibitem[\protect\citeauthoryear{Frey and Dueck}{2007}]{frey2007clustering}
Frey, B.~J., and Dueck, D.
\newblock 2007.
\newblock Clustering by passing messages between data points.
\newblock {\em science} 315(5814):972--976.

\bibitem[\protect\citeauthoryear{gui}{2010}]{guidecs2010}
2010.
\newblock Guide to diagnosis and treatment of cervical spondylosis.
\newblock Technical report, Chinese association of rehabilitation medicine.

\bibitem[\protect\citeauthoryear{Hoadley \bgroup et al\mbox.\egroup
  }{2014}]{hoadley2014multiplatform}
Hoadley, K.~A.; Yau, C.; Wolf, D.~M.; Cherniack, A.~D.; Tamborero, D.; Ng, S.;
  Leiserson, M.~D.; Niu, B.; McLellan, M.~D.; Uzunangelov, V.; et~al.
\newblock 2014.
\newblock Multiplatform analysis of 12 cancer types reveals molecular
  classification within and across tissues of origin.
\newblock {\em Cell} 158(4):929--944.

\bibitem[\protect\citeauthoryear{Hogg-Johnson \bgroup et al\mbox.\egroup
  }{2009}]{hogg2009burden}
Hogg-Johnson, S.; van~der Velde, G.; Carroll, L.~J.; Holm, L.~W.; Cassidy,
  J.~D.; Guzman, J.; C{\^o}t{\'e}, P.; Haldeman, S.; Ammendolia, C.; Carragee,
  E.; et~al.
\newblock 2009.
\newblock The burden and determinants of neck pain in the general population:
  results of the bone and joint decade 2000--2010 task force on neck pain and
  its associated disorders.
\newblock {\em Journal of manipulative and physiological therapeutics}
  32(2):S46--S60.

\bibitem[\protect\citeauthoryear{Howard~Vernon}{}]{howardneck}
Howard~Vernon, D.
\newblock The neck disability index.

\bibitem[\protect\citeauthoryear{Huang \bgroup et al\mbox.\egroup
  }{2018}]{huang2018clustering}
Huang, Y.; Zhan, J.; Wang, N.; Luo, C.; Wang, L.; and Ren, R.
\newblock 2018.
\newblock Clustering residential electricity load curves via community
  detection in network.
\newblock {\em arXiv preprint arXiv:1811.10356}.

\bibitem[\protect\citeauthoryear{Jebri \bgroup et al\mbox.\egroup
  }{2015}]{jebri2015detection}
Jebri, B.; Phillips, M.; Knapp, K.; Appelboam, A.; Reuben, A.; and Slabaugh, G.
\newblock 2015.
\newblock Detection of degenerative change in lateral projection cervical spine
  x-ray images.
\newblock In {\em Medical Imaging 2015: Computer-Aided Diagnosis}, volume 9414,
   941404.
\newblock International Society for Optics and Photonics.

\bibitem[\protect\citeauthoryear{Johnston \bgroup et al\mbox.\egroup
  }{2008a}]{johnston2008alterations}
Johnston, V., .; Jull, G., .; Darnell, R., .; Jimmieson, N.~L.; and Souvlis,
  T., .
\newblock 2008a.
\newblock Alterations in cervical muscle activity in functional and stressful
  tasks in female office workers with neck pain.
\newblock {\em European Journal of Applied Physiology} 103(3):253--264.

\bibitem[\protect\citeauthoryear{Johnston \bgroup et al\mbox.\egroup
  }{2008b}]{johnston2008neck}
Johnston, V., .; Jull, G., .; Souvlis, T., .; and Jimmieson, N.~L.
\newblock 2008b.
\newblock Neck movement and muscle activity characteristics in female office
  workers with neck pain.
\newblock {\em Spine} 33(5):555.

\bibitem[\protect\citeauthoryear{Junlin~Chen}{2017}]{chenjunlin2017}
Junlin~Chen, Runmin~Peng, S. L. X.~C.
\newblock 2017.
\newblock Self-organizing feature map neural netork and k-means algorithm as a
  data excavation tool for obtaining geological information from regional
  geochemical exploration data.
\newblock {\em Geophyical and Geochemical Exploration} 41(5):919--927.

\bibitem[\protect\citeauthoryear{Kotil and Bilge}{2008}]{kotil2008prospective}
Kotil, K., and Bilge, T.
\newblock 2008.
\newblock Prospective study of anterior cervical microforaminotomy for cervical
  radiculopathy.
\newblock {\em Journal of Clinical Neuroscience} 15(7):749--756.

\bibitem[\protect\citeauthoryear{Li \bgroup et al\mbox.\egroup
  }{2018}]{li2018feature}
Li, J.; Cheng, K.; Wang, S.; Morstatter, F.; Trevino, R.~P.; Tang, J.; and Liu,
  H.
\newblock 2018.
\newblock Feature selection: A data perspective.
\newblock {\em ACM Computing Surveys (CSUR)} 50(6):94.

\bibitem[\protect\citeauthoryear{MacQueen and others}{1967}]{macqueen1967some}
MacQueen, J., et~al.
\newblock 1967.
\newblock Some methods for classification and analysis of multivariate
  observations.
\newblock In {\em Proceedings of the fifth Berkeley symposium on mathematical
  statistics and probability}, volume~1,  281--297.
\newblock Oakland, CA, USA.

\bibitem[\protect\citeauthoryear{Madeleine \bgroup et al\mbox.\egroup
  }{2016}]{madeleine2016effects}
Madeleine, P.; Xie, Y.; Szeto, G. P.~Y.; and Samani, A.
\newblock 2016.
\newblock Effects of chronic neck-shoulder pain on normalized mutual
  information analysis of surface electromyography during functional tasks.
\newblock {\em Clinical Neurophysiology} 127(9):3110--3117.

\bibitem[\protect\citeauthoryear{Matz \bgroup et al\mbox.\egroup
  }{2009}]{matz2009joint}
Matz, P.; Anderson, P.; Holly, L.; Groff, M.; Heary, R.; Kaiser, M.; Mummaneni,
  P.; Ryken, T.; Choudhri, T.; Vresilovic, E.; et~al.
\newblock 2009.
\newblock Joint section on disorders of the spine and peripheral nerves of the
  american association of neurological surgeons and congress of neurological
  surgeons.
\newblock {\em J Neurosurg Spine} 11(2):157--169.

\bibitem[\protect\citeauthoryear{Nana~Wang}{}]{wang2018}
Nana~Wang, Xi~Huang, Y. R. J. X. J. L. N. W. L.~C.
\newblock A convenient non-harm cervical spondylosis intelligent identity
  method based on machine learning.
\newblock {\em Scientific Reports}.

\bibitem[\protect\citeauthoryear{Reynolds}{2015}]{reynolds2015gaussian}
Reynolds, D.
\newblock 2015.
\newblock Gaussian mixture models.
\newblock {\em Encyclopedia of biometrics}  827--832.

\bibitem[\protect\citeauthoryear{Roweis and Saul}{2000}]{roweis2000nonlinear}
Roweis, S.~T., and Saul, L.~K.
\newblock 2000.
\newblock Nonlinear dimensionality reduction by locally linear embedding.
\newblock {\em science} 290(5500):2323--2326.

\bibitem[\protect\citeauthoryear{Saxena \bgroup et al\mbox.\egroup
  }{2017}]{saxena2017review}
Saxena, A.; Prasad, M.; Gupta, A.; Bharill, N.; Patel, O.~P.; Tiwari, A.; Er,
  M.~J.; Ding, W.; and Lin, C.-T.
\newblock 2017.
\newblock A review of clustering techniques and developments.
\newblock {\em Neurocomputing} 267:664--681.

\bibitem[\protect\citeauthoryear{scikit-learn developers}]{metric}
scikit-learn developers.
\newblock calinski harabaz score.
\newblock
  \url{https://blog.csdn.net/u010159842/article/details/78624135}.
\newblock Accessed December 28, 2018.

\bibitem[\protect\citeauthoryear{Shen \bgroup et al\mbox.\egroup
  }{2012}]{shen2012integrative}
Shen, R.; Mo, Q.; Schultz, N.; Seshan, V.~E.; Olshen, A.~B.; Huse, J.; Ladanyi,
  M.; and Sander, C.
\newblock 2012.
\newblock Integrative subtype discovery in glioblastoma using icluster.
\newblock {\em PloS one} 7(4):e35236.

\bibitem[\protect\citeauthoryear{Shen, Olshen, and
  Ladanyi}{2009}]{shen2009integrative}
Shen, R.; Olshen, A.~B.; and Ladanyi, M.
\newblock 2009.
\newblock Integrative clustering of multiple genomic data types using a joint
  latent variable model with application to breast and lung cancer subtype
  analysis.
\newblock {\em Bioinformatics} 25(22):2906--2912.

\bibitem[\protect\citeauthoryear{Vernon and Mior}{1991}]{vernon1991neck}
Vernon, H., and Mior, S.
\newblock 1991.
\newblock The neck disability index: a study of reliability and validity.
\newblock {\em Journal of manipulative and physiological therapeutics}
  14(7):409--415.

\bibitem[\protect\citeauthoryear{Vesanto, Alhoniemi, and
  others}{2000}]{vesanto2000clustering}
Vesanto, J.; Alhoniemi, E.; et~al.
\newblock 2000.
\newblock Clustering of the self-organizing map.
\newblock {\em IEEE Transactions on neural networks} 11(3):586--600.

\bibitem[\protect\citeauthoryear{Von~Luxburg}{2007}]{von2007tutorial}
Von~Luxburg, U.
\newblock 2007.
\newblock A tutorial on spectral clustering.
\newblock {\em Statistics and computing} 17(4):395--416.

\bibitem[\protect\citeauthoryear{Wang \bgroup et al\mbox.\egroup
  }{2018}]{wang2018easicsdeep}
Wang, N.; Cui, L.; Huang, X.; Xiang, Y.; and Xiao, J.
\newblock 2018.
\newblock Easicsdeep: A deep learning model for cervical spondylosis
  identification using surface electromyography signal.
\newblock {\em arXiv preprint arXiv:1812.04912}.

\bibitem[\protect\citeauthoryear{Yu \bgroup et al\mbox.\egroup
  }{2015}]{yu2015classifying}
Yu, X.; Liu, M.; Meng, L.; and Xiang, L.
\newblock 2015.
\newblock Classifying cervical spondylosis based on x-ray quantitative
  diagnosis.
\newblock {\em Neurocomputing} 165:222--227.

\bibitem[\protect\citeauthoryear{Zhang, Ramakrishnan, and
  Livny}{1996}]{zhang1996birch}
Zhang, T.; Ramakrishnan, R.; and Livny, M.
\newblock 1996.
\newblock Birch: an efficient data clustering method for very large databases.
\newblock In {\em ACM Sigmod Record}, volume~25,  103--114.
\newblock ACM.

\bibitem[\protect\citeauthoryear{Zhongmin}{2011}]{CS2011zhaozhongmin}
Zhongmin, Z.
\newblock 2011.
\newblock {\em Study on the Symptoms and Soft Tissue Changes of Cervical
  Cervical Spondylosis}.
\newblock Ph.D. Dissertation, China Academy of Chinese Medical Sciences.

\end{thebibliography}
\end{document}